\documentclass{article}

\usepackage[preprint]{neurips_2022}

\usepackage{algpseudocode}
\usepackage{subcaption}
\usepackage{algorithm}
\usepackage{mathtools}
\usepackage{csquotes}
\usepackage{graphicx}
\usepackage{makecell}
\usepackage{multirow}
\usepackage{multicol}
\usepackage{amsmath}
\usepackage{amssymb}
\usepackage{amsthm}

\usepackage[utf8]{inputenc} 
\usepackage[T1]{fontenc}    
\usepackage{hyperref}       
\usepackage{url}            
\usepackage{booktabs}       
\usepackage{amsfonts}       
\usepackage{nicefrac}       
\usepackage{microtype}      
\usepackage{xcolor}         

\newcommand{\RR}{\mathbb{R}}
\newcommand{\PP}{\mathbb{P}}

\newcommand{\NN}{\mathcal{N}}

\newcommand{\EE}{\mathbb{E}}

\DeclareMathOperator{\var}{var}

\newtheorem{theorem}{Theorem}
\newtheorem{prop}[theorem]{Proposition}

\title{{Mitigating Out-of-Distribution Data Density Overestimation in Energy-Based Models}}

\author{
Beomsu Kim \\
Department of Mathematical Sciences \\
KAIST \\
\texttt{beomsu.kim@kaist.ac.kr} \\
\And
Jong Chul Ye \\
Graduate School of AI \\
KAIST \\
\texttt{jong.ye@kaist.ac.kr} \\
}

\begin{document}

\maketitle

\begin{abstract}
Deep energy-based models (EBMs),  which use deep neural networks (DNNs) as energy functions, are receiving increasing attention due to their ability to learn complex distributions. To train deep EBMs, the maximum likelihood estimation (MLE) with short-run Langevin Monte Carlo (LMC) is often used. While the MLE with short-run LMC is computationally efficient compared to an MLE with full Markov Chain Monte Carlo (MCMC), it often  assigns high density to out-of-distribution (OOD) data. To address this issue, here we systematically investigate why the MLE with short-run LMC can converge to EBMs with wrong density estimates, and reveal that the heuristic modifications to LMC introduced by previous works were the main problem. We then propose a Uniform Support Partitioning (USP) scheme that optimizes a set of points to evenly partition the support of the EBM and then uses the resulting points to approximate the EBM-MLE loss gradient. We empirically demonstrate that USP avoids the pitfalls of short-run LMC, leading to significantly improved OOD data detection performance on Fashion-MNIST.
\end{abstract}

\section{Introduction} \label{sec:intro}

For unsupervised learning, it is often of great interest to approximate a given data distribution using a generative model. Applications of generative models are abundant, ranging from data generation \cite{kingma2013,goodfellow2014} to out-of-distribution (OOD) data detection \cite{choi2018,nalisnick2019,ren2019,hendrycks2019,serra2020}, improving calibration and robustness of classifiers \cite{du2019}, etc. Among the wide variety of generative models, Energy-Based Models (EBMs) \cite{lecun2006} parametrized by deep neural networks (DNNs) have recently gained attention thanks to their flexibility in modeling complex distributions.

There are multiple ways of training EBMs, and the two most studied methods are maximum likelihood estimation (MLE) with Markov Chain Monte Carlo (MCMC) and Score Matching (SM) \cite{song2021}. Both methods have undergone appropriate modifications for training deep EBMs, i.e., EBMs parametrized by DNNs. For instance, in the case of SM, Song et. al \cite{song2019} proposed the estimation of gradients, not density, of the data distribution to bypass calculation of the Hessian. For MLE with MCMC, Du and Mordatch \cite{du2019} replaced MCMC, which often requires thousands of iterations until convergence, with short-run Langevin Monte Carlo (LMC) and a replay buffer.

Despite such developments, EBMs suffer from the problem of density overestimation on OOD data \cite{elflein2021}. Concretely, given an EBM trained by an MLE with short-run LMC (SRLMC), OOD data often have density values similar to or higher than that of training data. This does not make sense, since by the definition of OOD data, the supports of training data distribution and OOD data distribution do not intersect. Mahmood et. al \cite{mahmood2021} attempted to use score functions to detect OOD data, but they provided little insights into the density overestimation problem, since score functions model the gradient, not the density.

In this paper, we approach this problem in two ways. First, we rigorously investigate why and how the MLE with SRLMC can yield EBMs with wrong density estimates. Based on the observations, we then propose a novel technique, called Uniform Support Partitioning (USP), to solve the MLE for EBMs. In contrast to LMC which uses a stochastic process to sample from the EBM, USP solves a deterministic optimization problem to find points which uniformly partition the support of the EBM. USP then uses those points to approximate the MLE objective gradient through numerical integration.

We also introduce a practical version of USP, called Persistent Stochastic USP (PS-USP) and demonstrate on the problem of learning a mixture of Gaussians that PS-USP is capable of crossing low-density regions. On the Fashion-MNIST dataset, we show deep EBM trained with PS-USP shows significantly better OOD data detection performance than deep EBM trained with SRLMC.

Our contributions can be summarized as follows:
\begin{itemize}
\item Through theoretical analysis and experiments, we rigorously investigate why MLE with SRLMC could converge to an EBM with wrong density estimates,
and reveal that   it is caused by a combination of two heuristic modifications to LMC introduced by previous works: (a) early termination of LMC in short-run LMC and (b) using incorrect learning rate and noise scale ratio in LMC.
\item To avoid the pitfalls of MLE with SRLMC, we propose a novel technique, USP, to solve MLE for EBMs. USP solves an optimization problem to find a set of points which uniformly partition the support of EBMs. Then, it uses the points to approximate the MLE objective through numerical integration. We also introduce a practical version of USP for training deep EBMs.
\item We demonstrate on a toy example that USP is capable of accurately learning a distribution with multiple separated modes. We also show on the Fashion-MNIST dataset \cite{fmnist} that deep EBMs trained with USP attain significantly better OOD data detection performance than deep EBMs trained with SRLMC.
\end{itemize}

\section{Related Works}

\subsection{EBM Training via MLE with MCMC}

One of the most popular ways of training deep EBMs is maximizing the expected log-likelihood of the EBM via gradient ascent \cite{song2021}. However, calculating the  gradient of the log-likelihood of the EBM requires computing the expectation of the energy gradient on the current EBM distribution. A straightforward way to achieve this is to run MCMC on the EBM distribution and use the samples to approximate the energy gradient expectation. A popular choice of MCMC is Stochastic Gradient Langevin Dynamics (SGLD) \cite{welling2011}, a stochastic variant of LMC.

Recent works have taken further steps to make training deep EBMs efficient. Specifically, Du and Mordatch \cite{du2019} have proposed using non-convergent SRLMC (or short-run SGLD) with a replay buffer instead of LMC, which usually requires thousands of iterations until convergence, to sample from the EBM distribution. Latter works use the same technique as well \cite{nijkamp2019,grathwohl2020}. Furthermore, Yang et. al \cite{yang2021} combine short-run LMC with Pontryagin's Maximum Principle to reduce the number of forward and backward propagations.

The works by Nijkamp et. al \cite{nijkamp2019} and Nijkamp et. al \cite{nijkamp2020} perform an analysis of MLE with SRLMC and find that MLE with SRLMC trains EBMs to be data generators rather than density estimators. However, they do not explain how this leads to density overestimation for OOD data.

\subsection{OOD Data Detection with EBMs}

With the development of efficient deep EBM training methods, OOD data detection with EBMs also gained interest. Du and Mordatch \cite{du2019} discovered that an EBM trained by MLE with MCMC has slightly better OOD data detection performance than other deep density models such as Glow \cite{kingma2018} and PixelCNN++ \cite{salimans2017}. Grathwolh et. al \cite{grathwohl2020} incorporate label information into training EBMs and find that the EBM also outperforms Glow at OOD data detection. Finally, Elflein et. al \cite{elflein2021} discover that using supervision such as labels improves OOD detection on natural data and architectural modifications such as bottlenecks can also improve OOD detection.

\newpage

\section{Preliminaries}

Given an energy function $E_\theta : \RR^d \rightarrow \RR$ parametrized by $\theta$, an EBM is defined as
\begin{align}
q_\theta(x) = \frac{1}{Z(\theta)} \exp\{ -E_\theta(x) \}
\end{align}
where $Z(\theta)$ is the partition function, which ensures $q_\theta$ integrates to 1. Given a data distribution $p$, the EBM can be trained to approximate $p$ by MLE
\begin{align}
\max_\theta \EE_p [\log q_\theta(x)]
\end{align}
with gradient ascent. The gradient of the MLE objective can be decomposed into two terms:
\begin{align}
\nabla_\theta \EE_p [\log q_\theta(x)] = \EE_{q_\theta}[\nabla E_\theta(x)] - \EE_p[\nabla E_\theta(x)]. \label{eq:MLE_grad}
\end{align}
While we can easily calculate the second term since we have access to samples from $p$ (the training data), it is not the case for the first term. Previous works rely on LMC or its stochastic variant, SGLD, to sample from $q_\theta$ and calculate the first term. Given $x_0 \sim q_0(x)$ for some proposal distribution $q_0$, LMC iterates
\begin{align}
x_{t + 1} = x_t - \frac{\eta_t}{2} \nabla_x E_\theta(x) + \sqrt{\eta_t} \epsilon_t, \qquad t = 0, 1, 2, \ldots, T \label{eq:LMC}
\end{align}
where $\epsilon_t$ are i.i.d. standard normal Gaussian noises. For an appropriate choice of the sequence $\{\eta_t\}$, the sequence $\{x_t\}$ converges to a sample from $q_\theta$ as $T \rightarrow \infty$ \cite{welling2011,dalalyan2017}.

\subsection{MLE with Short-Run LMC (SRLMC)} \label{sec:mod_LMC}

A problem with MCMC is that it usually requires large number of iterations, i.e., large $T$ in \eqref{eq:LMC}, until convergence. This becomes problematic when we train deep EBMs, as forward and backward propagations of DNNs are expensive. In an attempt to alleviate this issue, Du and Mordatch \cite{du2019} propose three heuristic modifications to the EBM training procedure. These modifications have been adopted by latter works \cite{nijkamp2019,grathwohl2020,yang2021} for training EBMs as well. We now describe the modifications.

\textbf{Short-run LMC (SRLMC).} The first modification is using extremely small $T$. While conventional LMC can require thousands of iterations until convergence, Du and Mordatch \cite{du2019} propose using $T \leq 100$. Then, SRLMC samples are used to calculate the first term in \eqref{eq:MLE_grad}.

\textbf{Decoupling step size and noise scale.} The second modification is decoupling the gradient coefficient and noise coefficient in \eqref{eq:LMC}:
\begin{align}
x_{t + 1} = x_t - \frac{\alpha_t}{2} \nabla_x E_\theta(x) + \sqrt{\beta_t} \epsilon_t, \qquad t = 0, 1, 2, \ldots, T \label{eq:LMC_mod}
\end{align}
where $\alpha_t$ is called the \textit{step size} and $\beta_t$ is called the \textit{noise scale}. Since $T$ is set to be small, Du and Mordatch \cite{du2019} set $\alpha_t \gg \beta_t$ to accelerate the convergence of SRLMC.

\textbf{Replay buffer.} The third modification is to maintain a replay buffer of SRLMC samples. Specifically, instead of using random noise to initialize SRLMC at each iteration of EBM update, Du and Mordatch \cite{du2019} maintain a replay buffer which stores past SRLMC samples. A mixture of replay buffer samples and random noise is used to initialize SRLMC at each iteration of EBM training, and the outputs are used to update the replay buffer.

In the next section, we demonstrate that the first and second modifications can lead to EBMs with incorrect density, and the third modification does not alleviate the issue.

\begin{figure}[t!]
\centering
\begin{subfigure}{\linewidth}
\includegraphics[width=1.0\linewidth]{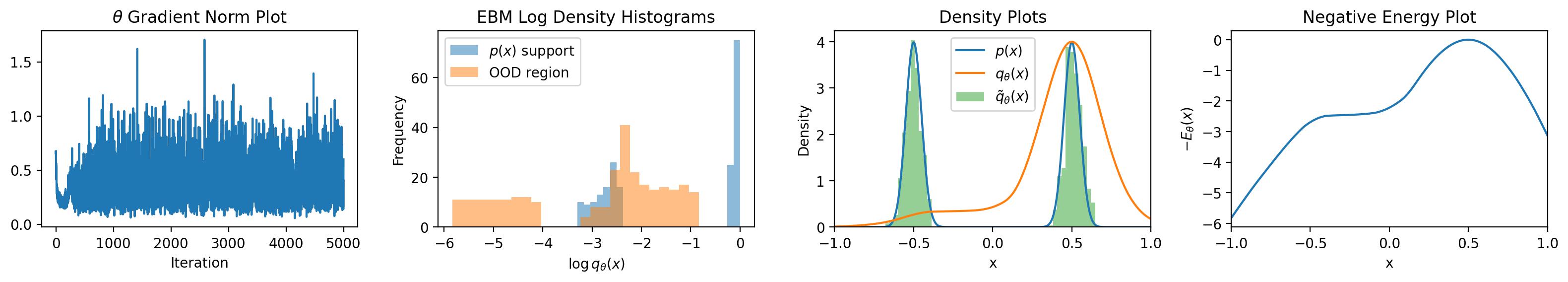}
\caption{Training via SRLMC with no replay buffer.}
\label{fig:sr_lmc_no_buf}
\end{subfigure}
\begin{subfigure}{\linewidth}
\includegraphics[width=1.0\linewidth]{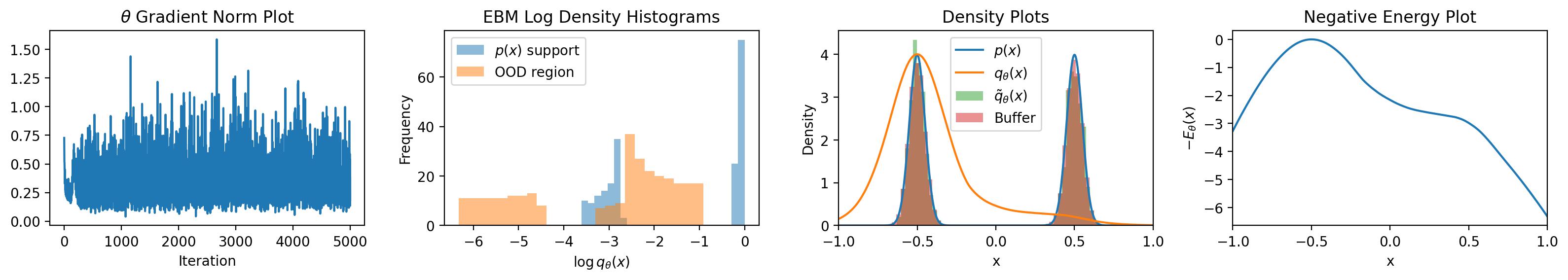}
\caption{Training via SRLMC with replay buffer.}
\label{fig:sr_lmc_buf}
\end{subfigure}
\begin{subfigure}{\linewidth}
\includegraphics[width=1.0\linewidth]{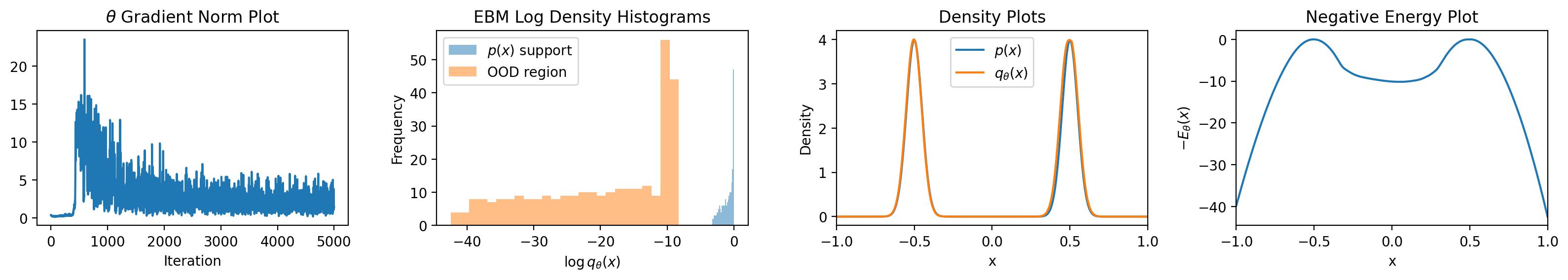}
\caption{Training via self-normalized importance sampling / Riemann sum.}
\label{fig:is}
\end{subfigure}
\caption{Comparison of methods for solving EBM MLE. The first column shows the evolution of $\theta$ gradient norm throughout the training process. The second column displays the trained EBM log density on the data distribution support and OOD regions. The third column shows densities and histograms of relevant distributions. The final column plots the negative energy for the trained EBM. Note that the density plots for $q_\theta(x)$ in (a) and (b) are unnormalized.}
\label{fig:mle_comp}
\end{figure}

\section{A Solution to MLE With SRLMC Can Overestimate OOD Data Density} \label{sec:overestimate}

From here on, we will refer to the EBM training procedure described in Section \ref{sec:mod_LMC} as MLE with SRLMC. For the moment, let us assume we do not use a replay buffer. Let $q_0$ be some proposal distribution, e.g., a uniform distribution, and let $\tilde{q}_\theta$ be the distribution of $\tilde{x}$ produced by running SRLMC on $x \sim q_0$\footnote{{We remark $\tilde{q}_\theta$ will generally not be equal to $q_\theta$ since SRLMC does not run LMC until convergence.}}. Then, EBM gradient update with MLE with SRLMC becomes
\begin{align}
\theta \leftarrow \theta + \EE_{\tilde{q}_\theta}[\nabla_\theta E_\theta(x)] - \EE_p[\nabla_\theta E_\theta(x)] \label{eq:update_mod}
\end{align}
and thus stationarity is achieved when
\begin{align}
\tilde{q}_\theta = p \label{eq:stationary}
\end{align}
for then the expectations in \eqref{eq:update_mod} will cancel out and no update to $\theta$ is made. We now demonstrate that a stationary point of \eqref{eq:update_mod} can assign high density to OOD regions. To this end, we consider the problem of training a deep EBM to approximate the mixture of two Gaussians $\NN(-0.5,0.05^2)$ and $\NN(0.5,0.05^2)$. The Gaussians are given equal weights. Also, $\alpha_t  = 10 \beta_t$ and $T = 40$.

Figure \ref{fig:sr_lmc_no_buf} shows the result of training an EBM with MLE with SRLMC and no replay buffer. The proposal distribution $q_0$ is the uniform distribution on $(-1,1)$. The leftmost $\theta$ gradient norm plot indeed shows the EBM has nearly converged to some stationary point\footnote{The gradient norm does not become exactly zero due to the stochasticity in LMC and finite number of samples used to approximate expectations. This causes oscillation of the gradient norm.}. However, contrary to our hopes, the second figure shows that a significant portion of the OOD region is assigned higher density than half of the data distribution support. The third and fourth figures indicate this is because the EBM has learned a density with wide modes at $x = \pm 0.5$, and the second mode is much lower than the first mode. 
Yet, despite the discrepancy between the EBM density and the data density, $\tilde{q}_\theta$ is identical to $p$. In particular, $\tilde{q}_\theta$ has two modes of equal height although the heights of modes of the EBM differ significantly.

So far, we have experimentally shown the existence of a stationary point of \eqref{eq:update_mod} which exhibits density overestimation and that a deep EBM can  converge to this undesirable point. We now explain why the EBM of Figure \ref{fig:sr_lmc_no_buf} is a stationary point of \eqref{eq:update_mod}. 
According to \eqref{eq:stationary}, this amounts to explaining how SRLMC can generate training data with this wrong EBM. Two factors play a role: poor mixing of SRLMC and incorrect step size and noise scale ratio.

\textbf{Poor mixing of SRLMC.} In general, LMC itself mixes very slowly. This issue was previously pointed out by Song and Ermon \cite{song2019}. As LMC uses the gradient information, it will initially tend to follow the steepest path of ascent of $-E_\theta(x)$. Theoretically, LMC \eqref{eq:LMC} will converge in the limit $T \rightarrow \infty$, but SRLMC terminates with a very small $T$. So, an SRLMC sample will typically end up in the mode whose basin of attraction contained its initialization point.

\begin{figure}[t!]
\centering
\begin{subfigure}{0.314\linewidth}
\includegraphics[width=1.0\linewidth]{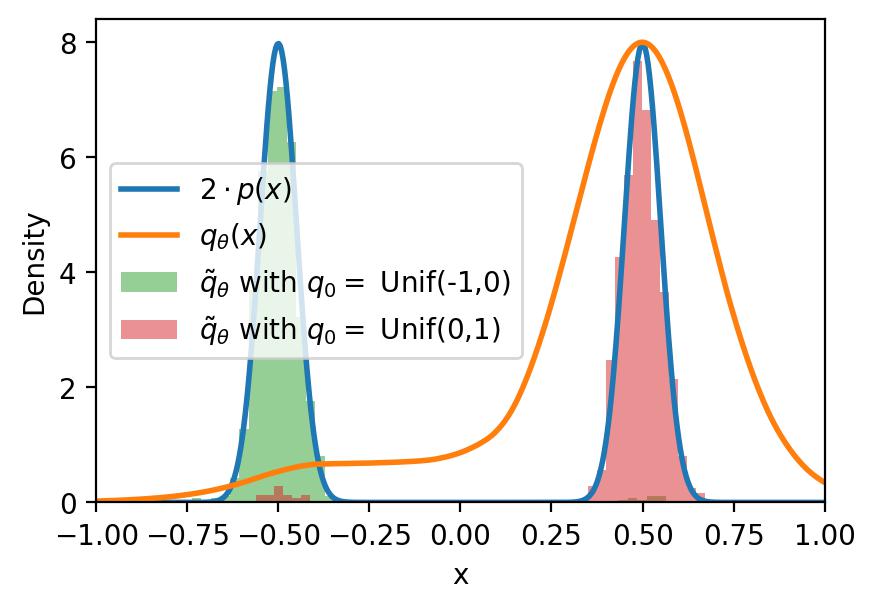}
\end{subfigure}
\begin{subfigure}{0.32\linewidth}
\includegraphics[width=1.0\linewidth]{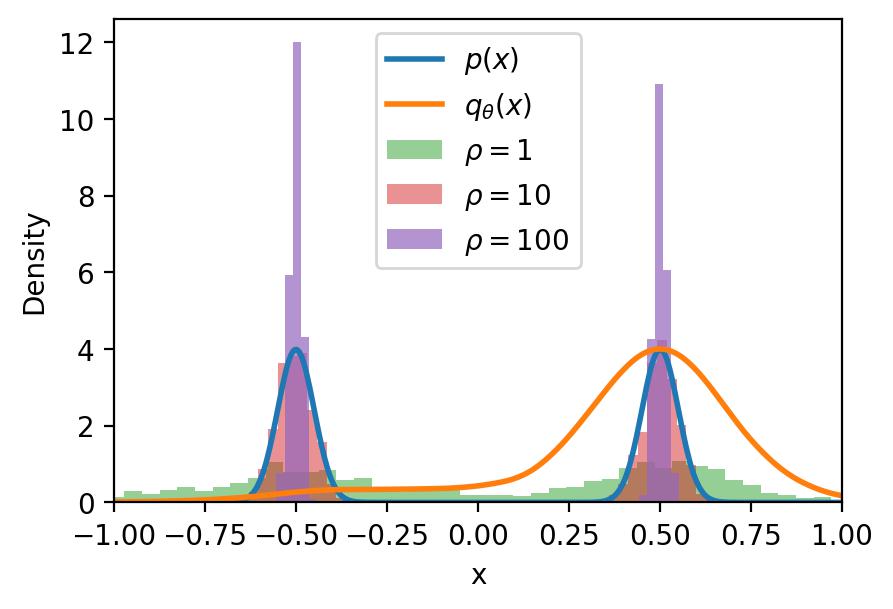}
\end{subfigure}
\begin{subfigure}{0.32\linewidth}
\includegraphics[width=1.0\linewidth]{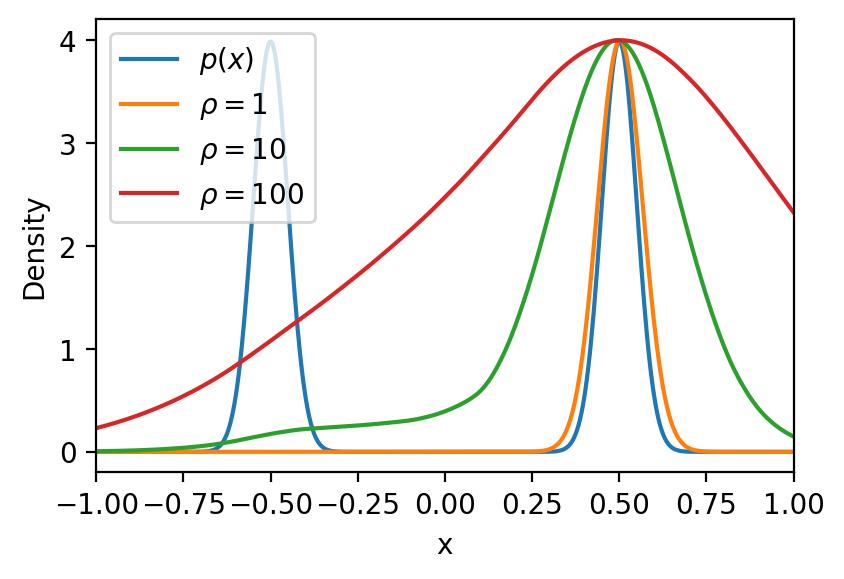}
\end{subfigure}
\caption{Effect of SRLMC. \textbf{Left:} histograms of SRLMC samples on the EBM of Figure \ref{fig:sr_lmc_no_buf} with $q_0 = $ uniform distribution on $(-1,0)$ and $q_0 = $ uniform distribution on $(0,1)$. \textbf{Middle:} histograms of SRLMC samples with $q_0 = $ uniform distribution on $(-1,1)$ and various $\rho$. \textbf{Right:} EBMs trained to approximate $p$ via MLE with SRLMC with various $\rho$. We remark that for each EBM, $\tilde{q}_\theta = p$ when the same $\rho$ is used for training and sample generation. Also, each EBM has a slight bump, i.e, a mode, at $x = -0.5$, so SRLMC samples form the mode at $x = -0.5$.}
\label{fig:srlmc_analysis}
\end{figure}

{The EBM can exploit this pathology of SRLMC and learn a density with modes of incorrect probability mass. Specifically, let us consider a mode of an EBM, denoted $m_q$, and its basin of attraction $B_q$. 
We denote the probability mass contained in the corresponding data mode $m_p$ as $M_p$. Now, due to poor mixing of SRLMC, most points initialized in $B_q$ will end up in $m_q$. So, if we denote the probability measure of $q_0$ as $\mathbb{Q}_0$, the ratio of proposal samples that will be placed in $m_q$ by SRLMC will be approximately $\mathbb{Q}_0(B_q)$. Since the EBM learns to match the distribution of generated data with training data (c.f. Eq. \eqref{eq:stationary}), at convergence of the EBM, we must have $\mathbb{Q}_0(B_q) \approx M_p$. As $B_q$ is generally unrelated to the probability mass of $m_q$, the modes of EBM can have incorrect probability mass. The left panel of Figure \ref{fig:srlmc_analysis} confirms this: SRLMC samples initialized on $(-1,0)$ mostly converge to the left mode, and samples initialized on $(1,0)$ mostly converge to the right mode.}


The above observations have serious implications in high dimensions. In high dimensions, samples from $q_0$ can come from a very small subset of the support of $q_0$. As the following proposition shows, this phenomenon holds for a wide variety of $q_0$.

\begin{prop} \label{prop:1}
Suppose $X$ is a $d$-dimensional random vector whose components are i.i.d. with mean $\mu$, variance $\sigma^2$, and finite fourth moment. Then, for any $\epsilon > 0$,
\begin{align*}
\lim_{d \rightarrow \infty} \PP\left\{(1 - \epsilon) \sqrt{d(\sigma^2 + \mu^2)} < \|X\|_2 < (1 + \epsilon) \sqrt{d(\sigma^2 + \mu^2)}\right\} = 1.
\end{align*}
\end{prop}

Proposition \ref{prop:1} claims that if the components of a high-dimensional vector is i.i.d. with finite variance, almost all samples come from a thin shell. This result is applicable to high-dimensional Gaussian distributions\footnote{If the Gaussian distribution has a non-identity covariance matrix, Proposition \ref{prop:1} can be easily extended to show that samples will lie around the boundary of an ellipsoid with high probability.} and uniform distributions on $[-1,1]^d$, which are common choices of $q_0$ \cite{nijkamp2019,grathwohl2020,yang2021}.

Let us call this thin shell $S$. Then, just like the case of Figure \ref{fig:sr_lmc_no_buf}, the EBM can learn a path of ascent from $S$ to the data support. Moreover, the EBM is free to assign arbitrary density to OOD regions which do not intersect $S$, the data support, and the path of ascent. Since $S$ is very small and natural data lies on low-dimensional manifolds (if we adopt the manifold hypothesis), the volume of such OOD region can be very large in high dimensions. The poor OOD data detection performance of EBMs trained with SRLMC observed by Elflein et. al \cite{elflein2021} provides evidence for this claim.

\textbf{Incorrect step size and noise scale ratio.} The first factor,  poor mixing of SRLMC, explains how the wrong EBM in Fig \ref{fig:sr_lmc_no_buf} can generate two modes, each of which has same probability mass as the corresponding data mode. But, it does not explain how the EBM, which has wide modes, can generate data, which has narrow modes. Incorrect step size and noise scale ratio in \eqref{eq:LMC_mod} is to blame.

\begin{prop} \label{prop:2}
Assume the sequences $\{\alpha_t\}$ and $\{\beta_t\}$ in \eqref{eq:LMC_mod} satisfy $\alpha_t / \beta_t = \rho$ for some $\rho > 0$ for all $t$. Also, assume the sequence generated by \eqref{eq:LMC_mod} converges. Then $\{x_t\}$ converges in distribution to
\begin{align}
q_\theta^{\rho}(x) \coloneqq \frac{1}{Z(\theta, \rho)} \exp\left\{ - \rho E_\theta(x) \right\}.
\end{align}
\end{prop}

Proposition \ref{prop:2} tells us that $\rho$ controls the sharpness of the sampled distribution. LMC with large $\rho$ will sample from a sharpened version of $q_\theta$, and LMC with small $\rho$ will sample from a wide version of $q_\theta$. The middle panel of Figure \ref{fig:srlmc_analysis} illustrates this fact. Together with the observation that SRLMC does not mix well, we can explain how the EBM of Figure \ref{fig:sr_lmc_no_buf} can generate training data with SRLMC.

Since SRLMC does not mix well, without loss of generality, we can focus on one mode of the EBM. As mentioned in Section \ref{sec:mod_LMC}, MLE with SRLMC uses large $\rho$ to accelerate convergence. So, SRLMC samples from a sharpened version of $q_\theta$. Hence, though the EBM has wide modes, $\tilde{q}_\theta$ has sharp modes which match the training data distribution. Furthermore, Proposition \ref{prop:2} can be used to prove:

\begin{prop} \label{prop:3}
Assume the EBM $q_\theta$ is trained via MLE with convergent modified LMC \eqref{eq:LMC_mod} with $\alpha_t / \beta_t = \rho > 0$. Then, $\theta$ such that
\begin{align}
q_\theta(x) \propto p(x)^{1/\rho}.
\end{align}
is a stationary point of MLE with gradient ascent.
\end{prop}

Indeed, the third panel of Figure \ref{fig:srlmc_analysis} shows smaller $\rho$ leads to sharper modes and larger $\rho$ leads to even wider modes.  So, combined with incorrect probability mass within modes due to poor mixing of SRLMC, larger $\rho$ exacerbates OOD data density overestimation.

\textbf{A replay buffer does not help.} Let us now consider the scenario where we use a replay buffer as well. Figure \ref{fig:sr_lmc_buf} shows that MLE with SRLMC and a buffer has converged to essentially the same solution (up to reflection w.r.t. the y-axis) as SRLMC without the buffer. This implies that the buffer does not alleviate OOD data overestimation.

\begin{figure}[t!]
\centering
\begin{subfigure}{0.51\linewidth}
\includegraphics[width=1.0\linewidth]{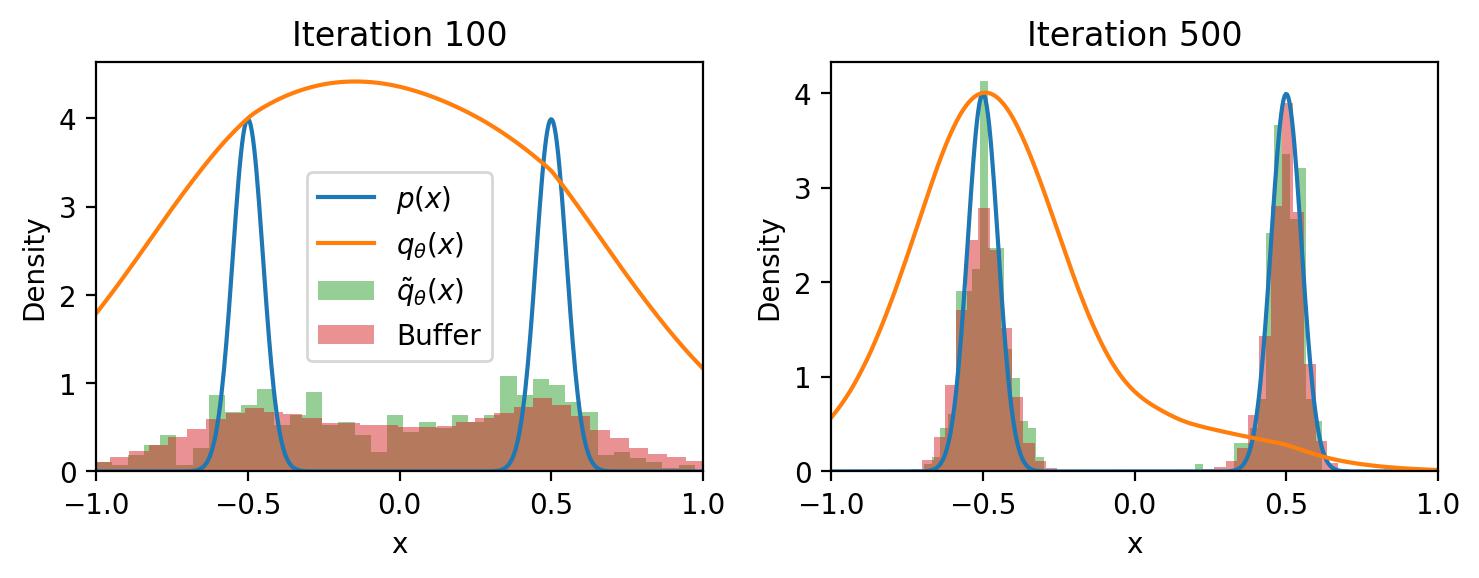}
\caption{Replay buffer evolution}
\label{fig:buf_evol}
\end{subfigure}
\hfill
\begin{subfigure}{0.47\linewidth}
\includegraphics[width=1.0\linewidth]{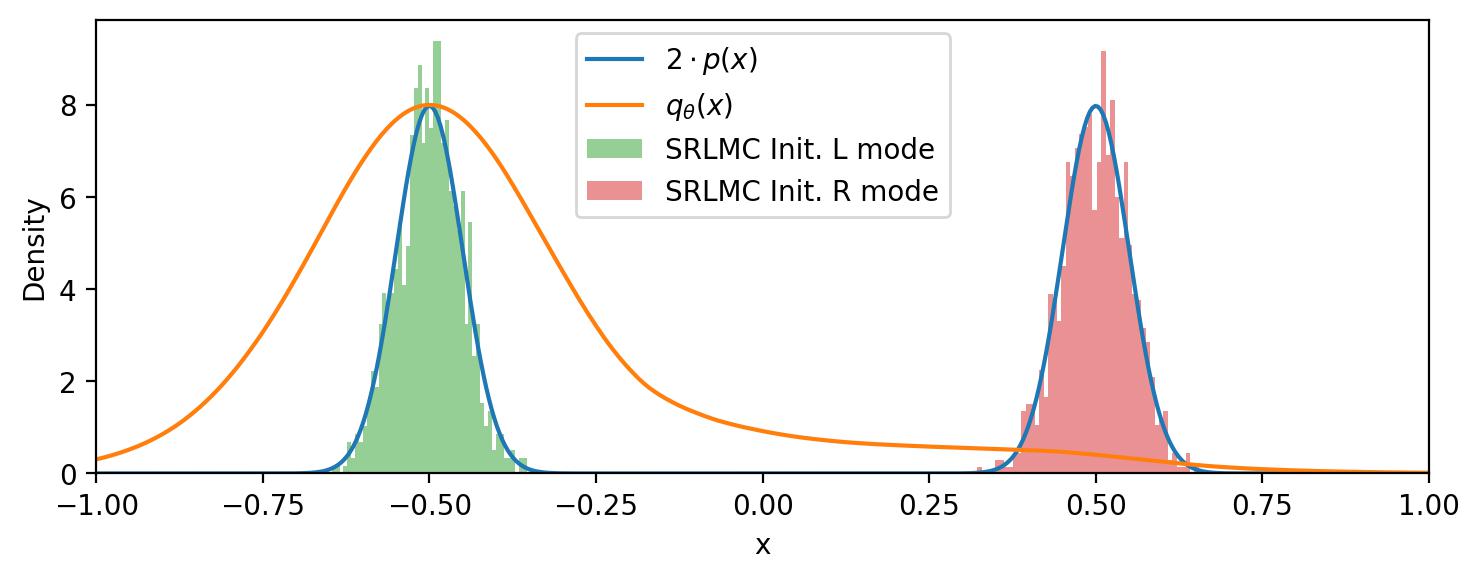}
\caption{SRLMC initialized from buffer (data)}
\label{fig:buf_srlmc}
\end{subfigure}
\caption{Analysis of MLE with SRLMC and replay buffer.}
\label{fig:srlmc_buf_analysis}
\end{figure}

{In fact, Figure \ref{fig:buf_evol} shows that as the EBM converges to the same solution as that of Figure \ref{fig:sr_lmc_buf}, the buffer sample distribution converges to the data distribution. Next, as illustrated in Figure \ref{fig:buf_srlmc}, even if SRLMC is initialized from buffer samples (which are now equal to data samples), SRLMC still does not mix well: SRLMC initialized from the left mode stays at the left mode, and SRLMC initialized from the right mode stays at the right mode. Thus, a replay buffer does not help mixing of SRLMC, so SRLMC with replay buffer suffers from the same problems as MLE with SRLMC without buffer.}

\subsection{A Simple Method that Avoids the Pitfalls of SRLMC on Low Dimensions} \label{sec:low_dim}

So far, we have shown the heuristic modifications of SRLMC in Section \ref{sec:mod_LMC} admits stationary points to the MLE problem which exhibit OOD data density overestimation. Several factors, poor mixing of SRLMC and incorrect step size and noise scale ratio, play a role in this. Furthermore, we have provided experimental and theoretical evidence that these factors could cause an EBM to converge to that problematic point. So, to train EBMs with correct density estimates, we need a method to estimate integral w.r.t. $q_\theta$ which is resilient to getting trapped in modes and accurate at estimating probability mass within each mode.

In low dimensions, there is a method which meets these desiderata: self-normalized importance sampling with the importance distribution as the uniform distribution on $\Omega$, or equivalently, Riemann sum as an approximation to the integral. Concretely, let us assume $q_\theta$ is supported on a subset of a compact domain $\Omega$. Then, given a set of points $\{u_i\}_{i = 1}^n$ which are uniformly sampled from $\Omega$ or form a partition of $\Omega$, we can approximate the expectation of a function $f$ with respect to $q_\theta$ as
\begin{align*}
\EE_{q_\theta}[f(x)] = \int_\Omega f(x) q_\theta(x) \, dx = \frac{\int_\Omega f(x) \exp\{-E_\theta(x)\} \, dx}{\int_\Omega \exp\{-E_\theta(x)\} \, dx} \approx \frac{\frac{1}{n} \sum_{i = 1}^n f(u_i) \exp\{-E_\theta(u_i)\}}{\frac{1}{n} \sum_{j = 1}^n \exp\{-E_\theta(u_j)\}}
\end{align*}
which can be concisely written as
\begin{align}
\EE_{q_\theta}[f(x)] \approx \sum_{i = 1}^n w_i f(u_i), \qquad w_i \coloneqq \frac{\exp\{-E_\theta(u_i)\}}{\sum_{j = 1}^n \exp\{-E_\theta(u_j)\}}. \label{eq:approx}
\end{align}

In the perspective of self-normalized importance sampling, the approximation \eqref{eq:approx} converges to the expectation as the size of uniform distribution samples $n \rightarrow \infty$. In the perspective of Riemann sum, the approximation converges to the true value as $n \rightarrow \infty$ and the norm of the partition, i.e., the maximum distance between two points in $\{u_i\}_{i = 1}^n$, converges to zero.

According to Figure \ref{fig:is}, using the MLE gradient \eqref{eq:MLE_grad} approximated by this method shows excellent performance on the problem of learning the mixture of two Gaussians. Contrary to SRLMC, this method is not affected by the fact that the two Gaussians have approximately disjoint support. So, the EBM trained by self-normalized importance sampling / Riemann sum places correct probability mass within each mode.

Unfortunately, in general, this method is not applicable to learning high-dimensional distributions. Suppose $p$ is the distribution of natural images, where we set $\Omega = [0,1]^d$. Also, assume $\{u_n\}_{i = 1}^n$ in \eqref{eq:approx} is distributed uniformly on $\Omega$, and $d$ is the dimension of images, where $d$ is generally very large. By Proposition \ref{prop:1}, most samples from the uniform distribution on $\Omega$ lie on a thin shell $S$. The volume of $S$ compared to the volume of $\Omega$ is vanishingly small, so $\{u_i\}_{i = 1}^n$ is unlikely to come from the high-density regions of $q_\theta$. Thus, the approximation \eqref{eq:approx} becomes increasingly inaccurate with larger $d$. Hence, in the next section, we propose a way to make self-normalized importance sampling / Riemann sum work in high dimensions.

\section{Uniform Support Partitioning (USP)}

To overcome the curse of dimensionality described in Section \ref{sec:low_dim}, we propose finding $\{u_i\}_{i = 1}^n$ which lie uniformly on the support of $q_\theta$. To this end, we solve the following optimization problem:
\begin{align}
\max_{u_i \in \Omega} \sum_{i = 1}^n \log q_\theta(u_i) \quad \text{subject to} \quad \|u_i - u_j\|_2 \geq \epsilon \ \text{for all} \ i \neq j. \label{eq:usp_opt}
\end{align}
Here, $\epsilon$ is a parameter which controls the fineness of the partition\footnote{If the Lebesgue measure of $\Omega$ is positive, the feasible set of \eqref{eq:usp_opt} will be nonempty for sufficiently small $\epsilon$.}. Intuitively, the above problem fills up the support of $q_\theta$ with $\epsilon$-balls centered at $\{u_i\}_{i = 1}^n$ with priority on high density regions. The points $\{u_i\}_{i = 1}^n$ are then used to approximate $\EE_{q_\theta}[\nabla_\theta E_\theta(x)]$ in the MLE gradient \eqref{eq:MLE_grad} using the formula \eqref{eq:approx}.

To solve \eqref{eq:usp_opt}, we take motivation from projected gradient ascent (PGA). Specifically, we iterate between a maximization step and a projection step. In the maximization step, we locally push each $u_i$ in the direction which maximizes the density. In the projection step, we perturb each $u_i$ such that the pairwise distance between points in $\{u_i\}_{i = 1}^n$ is $\geq \epsilon$. In the following paragraph, we give the full detail of our algorithm, which we call Uniform Support Partitioning (USP).

USP proceeds by iterating two steps. Suppose $\{u_i\}_{i = 1}^n$ is the set of points produced by USP in the previous iteration. The first step, called the \textit{maximization step}, seeks new points $u_i'$ in the proximity of $u_i$ which maximize the log-density. Specifically, we solve
\begin{align}
\qquad \max_{u_i' \in \Omega} \sum_{i = 1}^n \log q_\theta(u_i') \label{eq:usp_max}
\end{align}
via projected gradient ascent (PGA). The second step, called the \textit{repulsion step}, repels the points $\{u_i\}_{i = 1}^n$ apart so the constraint of \eqref{eq:usp_opt} is satisfied. Since there is no closed form formula for projecting an arbitrary set of points on the constraint set of \eqref{eq:usp_opt}, we use the gradient method to repel points from one another. Concretely, we solve
\begin{align}
\max_{u_i \in \Omega} \sum_{i \neq j} \min\{\|u_i - u_j\|_2, \epsilon\} \label{eq:usp_repul}
\end{align}
again via PGA. In practice, there is no guarantee that  particles $\{u_i\}_{i = 1}^n$ converge to the solution of \eqref{eq:usp_opt}, since we do not use an exact projection step. Nonetheless, as we will show in Section \ref{sec:exp}, we find this poses no problem in learning EBMs.


\subsection{Persistent Stochastic USP (PS-USP) for Training Deep EBMs} \label{sec:psusp}

If we are to train a deep EBM with USP, we face two problems: (a) if $n$ is large, each evaluation of the objective of \eqref{eq:usp_max} and \eqref{eq:usp_repul} can be expensive, and (b) running USP until convergence at each iteration of EBM gradient update can be computation costly. We introduce two modifications to USP which address these problems.

\textbf{Stochastic updates.} Suppose we wish to run USP on $\{u_i\}_{i = 1}^n$ where $n$ is very large. At each iteration of USP, we randomly choose $\Lambda \subsetneq [n]$ and optimize $\{u_i\}_{i \in \Lambda}$. The maximization step with $\{u_i\}_{i \in \Lambda}$ poses no problem, as the objective of \eqref{eq:usp_max} is separable. However, naively applying the repulsion step \eqref{eq:usp_repul} to only $\{u_i\}_{i \in \Lambda}$ can cause $\{u_i\}_{i \in \Lambda}$ to collapse into the same configuration as $\{u_j\}_{j \in [n] - \Lambda}$. So, to alleviate this issue, at each iteration of PGA of the repulsion step, we solve
\begin{align}
\max_{\{u_i\}_{i \in \Lambda} \in \Omega^{|\Lambda|}} \sum_{i \in \Lambda} \sum_{j \in \Lambda \cup \Gamma} 1_{i \neq j} \cdot \min\{\|u_i - u_j\|_2, \epsilon\}. \label{eq:usp_repul_stoc}
\end{align}
where $\Gamma \subset [n] - \Lambda$ is sampled uniformly at random.

\textbf{Persistent USP.} In the spirit of persistent contrastive divergence \cite{tieleman2008}, we only run a small number of USP iterations before each gradient update of the EBM. While we have no theoretical justification for this choice, experiments that follow show persistent USP performs sufficiently well.

USP with the above modifications is called Persistent Stochastic USP (PS-USP). One iteration of EBM parameter $\theta$ update with PS-USP proceeds as follows: (a) sample $\Lambda \subsetneq [n]$, (b) apply the maximization step to $\{u_i\}_{i \in \Lambda}$ via $n_m$ steps of PGA, (d) apply the repulsion step with \eqref{eq:usp_repul_stoc} with $n_r$ steps of PGA, (e) repeat steps (b) to (d) $N$ times, (e) choose $n_s$ points from $\{u_i\}_{i = 1}^n$ and calculate $\EE_{q_\theta}[\nabla_x E_\theta(x)]$ via \eqref{eq:approx}, (f) calculate the MLE gradient \eqref{eq:MLE_grad} and update $\theta$ via gradient ascent.

\begin{figure}[t!]
\centering
\begin{subfigure}{0.243\linewidth}
\includegraphics[width=1.0\linewidth]{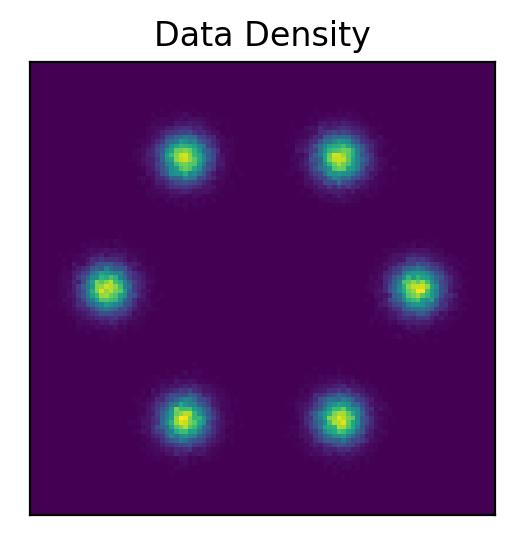}
\caption{MoG density}
\label{fig:data_mog}
\end{subfigure}
\hfill
\begin{subfigure}{0.238\linewidth}
\includegraphics[width=1.0\linewidth]{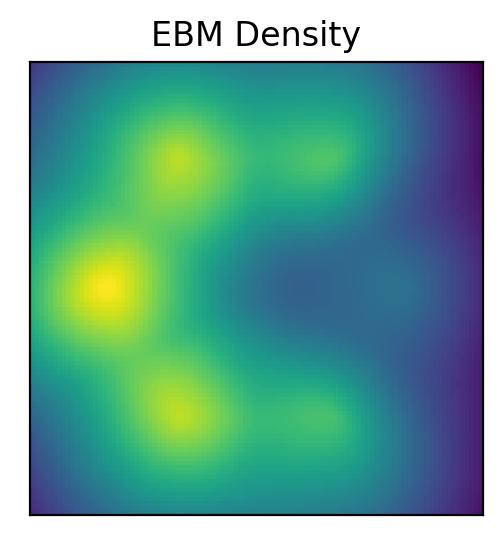}
\caption{SRLMC result}
\label{fig:srlmc_mog}
\end{subfigure}
\hfill
\begin{subfigure}{0.475\linewidth}
\includegraphics[width=1.0\linewidth]{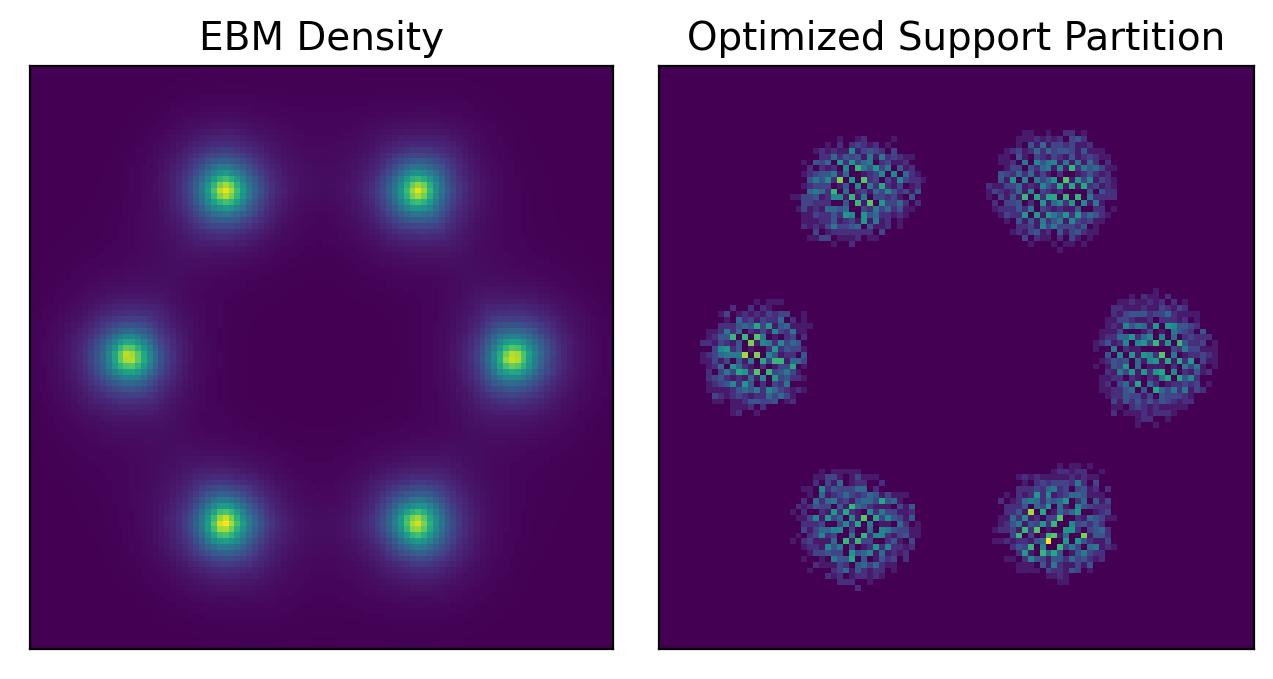}
\caption{PS-USP results}
\label{fig:usp_mog}
\end{subfigure}
\caption{Comparison of MLE with SRLMC and PS-USP on a 2D MoG.}
\label{fig:mog}
\end{figure}

\section{Numerical Experiments} \label{sec:exp}

\subsection{2 Dimensional Mixture of Gaussians (2D MoG)}

We first consider the problem of learning a deep EBM on a 2D mixture of Gaussians (MoG) with six modes (Figure \ref{fig:data_mog}). To check the robustness SRLMC and PS-USP to separated modes, for SRLMC, we set $q_0$ as the rightmost mode of the MoG, and for PS-USP, we initialize $\{u_i\}_{i = 1}^n$ as samples from the rightmost mode of the MoG. Each method was run until the EBM converged.

Figure \ref{fig:srlmc_mog} shows the result of using SRLMC to train a deep EBM to approximate the MoG. We again observe that the EBM exhibits incorrect probability mass ratio of the modes due to poor mixing of SRLMC. Also, the EBM density is blurry due to the wrong step size and noise scale ratio. Consequently, some OOD regions have higher density than the rightmost mode.

On the other hand, according to Figure \ref{fig:usp_mog}, the EBM trained by PS-USP has accurately learned the MoG. Moreover, the partition points are almost uniformly distributed on the support of the EBM. Thus, PS-USP mitigates the pitfalls of SRLMC .

\subsection{Fashion-MNIST}

{We now turn to the more challenging task of training deep EBMs on Fashion-MNIST \cite{fmnist} and using the EBM for OOD data detection. We use a simple CNN-based discriminative model as the energy function, without any special structure such as bottlenecks. After training, we evaluate the OOD data detection performances of EBMs using density values only. Evaluation metrics are false positive rate at true positive rate $95\%$ (FPR95) and the area under the precision-recall curve (AUPR). OOD data are MNIST \cite{mnist}, KMNIST \cite{kmnist}, NotMNIST \cite{nmnist}, Constant which consists of constant-valued images whose values are sampled from the uniform distribution on $[0,1]$, and Noise which consists of a mixture of uniform noise and standard Gaussian noise.}

\begin{figure}[t!]
\centering
\includegraphics[width=1.0\linewidth]{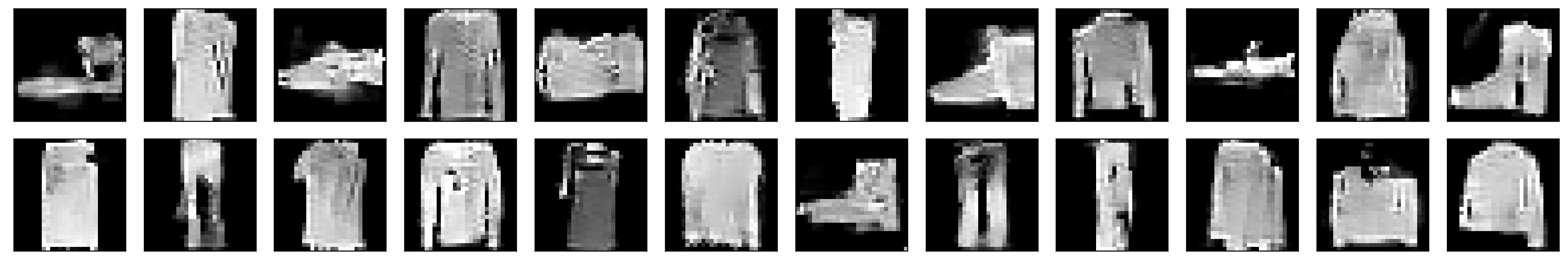}
\caption{An example of EBM support partition points found by PS-USP on FMNIST.}
\label{fig:fmnist_support}
\end{figure}

{Table \ref{table:fmnist_ood} compares the OOD data detection performances. We observe that except in the case of Noise, which is an easy OOD data to detect, EBMs trained with PS-USP beat EBMs trained with SRLMC by a nontrivial margin. In particular, FPR95 scores show significant gaps. This provides concrete evidence that USP can avoid the pitfalls of SRLMC in training deep EBMs. Also, Figure \ref{fig:fmnist_support} displays partition points found by PS-USP for the EBM trained by PS-USP. We observe that the partition points resemble actual FMNIST data, which implies the EBM has successfully learned to approximate the FMNIST distribution.}

\begin{table*}[t]
\centering
\resizebox{1.0\textwidth}{!}{
\begin{tabular}{c c c c c c c c c c c}
\toprule
OOD Data& \multicolumn{2}{c}{\textbf{MNIST}} & \multicolumn{2}{c}{\textbf{KMNIST}} & \multicolumn{2}{c}{\textbf{NotMNIST}} & \multicolumn{2}{c}{\textbf{Constant}} & \multicolumn{2}{c}{\textbf{Noise}} \\
\cmidrule(lr){1-1} \cmidrule(lr){2-3} \cmidrule(lr){4-5} \cmidrule(lr){6-7} \cmidrule(lr){8-9} \cmidrule(lr){10-11}
Statistic & FPR95 $\downarrow$ & AUPR $\uparrow$ & FPR95 $\downarrow$ & AUPR $\uparrow$ & FPR95 $\downarrow$ & AUPR $\uparrow$ & FPR95 $\downarrow$ & AUPR $\uparrow$ & FPR95 $\downarrow$ & AUPR $\uparrow$ \\
\cmidrule{1-11}
SRLMC  & $97.73_{\pm 0.76}$ & $86.01_{\pm 4.94}$ & $50.56_{\pm 6.02}$ & $90.57_{\pm 3.13}$ & $12.27_{\pm 3.06}$ & $99.05_{\pm 0.21}$ & $20.03_{\pm 12.47}$ & $99.89_{\pm 0.10}$ & $0.0_{\pm 0.0}$ & $100.0_{\pm 0.0}$ \\
PS-USP & $2.18_{\pm 1.03}$ & $99.15_{\pm 0.29}$ & $1.84_{\pm 1.01}$ & $99.32_{\pm 0.12}$ & $7.04_{\pm 0.57}$ & $99.07_{\pm 0.27}$ & $0.97_{\pm 0.78}$ & $99.99_{\pm 0.00}$ & $0.0_{\pm 0.0}$ & $100.0_{\pm 0.0}$ \\
\bottomrule
\end{tabular}}
\caption{Comparison of OOD data detection performances on FMNIST. $\downarrow$ indicates lower is better, and $\uparrow$ means higher is better. We report the mean and standard deviation over three trials.}
\label{table:fmnist_ood}
\end{table*}

\section{Limitations and Societal Impacts} \label{sec:limitations}

\textbf{Limitations.} The main limitation of this work is that we do not have convergence guarantees for USP and PS-USP. This is because we use approximate projection instead of exact projection in the repulsion step. Yet, we observed that USP performs well in practice. We believe a search for convergence proofs of USP could lead to better ways of solving \eqref{eq:usp_opt} and ultimately better algorithms for training deep EBMs. Thus, we believe this is a promising direction for future research.

\textbf{Societal impacts.} As EBMs are generative models, there is always a possibility that it may be exploited for malicious purposes, e.g., for producing Deepfakes or fake information. On the other hand, there are increasing efforts to apply EBMs in safety critical domains such as medical imaging. We believe our research is necessary for increasing the robustness of such machine learning systems, as our research concerns making deep generative models \enquote{know what they don't know} \cite{nalisnick2019b}.

\section{Conclusions}

In this work, we investigated why EBMs assign high density to OOD regions. We found that poor mixing of SRLMC and incorrect step size and noise scale ratio were the causes.  Motivated by these observations, we proposed a novel numerical integration method, USP which finds a uniform partition of the EBM support and uses the partition points to calculate the MLE gradient. We demonstrated on a MoG data that USP overcomes the pitfalls of SRLMC. Further, we showed that EBMs trained by USP has significantly better OOD data detection performance on FMNIST. We believe a theoretical analysis of USP could lead to better EBM training algorithms, and leave this for future work.

\newpage

{\small
\bibliographystyle{unsrt}
}


\newpage

\appendix

\section{Missing Proofs} \label{append:proofs}

\textbf{Proposition 1.} \textit{Suppose $X$ is a $d$-dimensional random vector whose components are i.i.d. with mean $\mu$, variance $\sigma^2$, and finite fourth moment. Then, for any $\epsilon > 0$,}
\begin{align*}
\lim_{d \rightarrow \infty} \PP\left\{(1 - \epsilon) \sqrt{d(\sigma^2 + \mu^2)} < \|X\|_2 < (1 + \epsilon) \sqrt{d(\sigma^2 + \mu^2)}\right\} = 1.
\end{align*}

\begin{proof}
We observe that
\begin{align}
\var(X_i^2) \leq \EE[X_i^4] < \infty
\end{align}
by assumption. So, by the $L^2$ weak law of large numbers,
\begin{align}
\frac{1}{d} \|X\|_2^2 = \frac{1}{d} \sum_{i = 1}^d X_i^2 \rightarrow \EE[X_i^2] = \sigma^2 + \mu^2
\end{align}
in probability as $d \rightarrow \infty$. This implies the claim of the proposition.
\end{proof}

\textbf{Proposition 2.} \textit{Assume the sequences $\{\alpha_t\}$ and $\{\beta_t\}$ in \eqref{eq:LMC_mod} satisfy $\alpha_t / \beta_t = \rho$ for some $\rho > 0$ for all $t$. Also, assume the sequence generated by \eqref{eq:LMC_mod} converges. Then $\{x_t\}$ converges in distribution to}
\begin{align*}
q_\theta^{\rho}(x) \coloneqq \frac{1}{Z(\theta, \rho)} \exp\left\{ - \rho E_\theta(x) \right\}.
\end{align*}

\begin{proof}
It is known that with the LMC iteration
\begin{align}
x_{t + 1} = x_t - \frac{\eta_t}{2} \nabla_x E_\theta(x) + \sqrt{\eta_t} \epsilon_t, \qquad t = 0, 1, 2, \ldots, T, \label{eq:prop2_1}
\end{align}
the sequence $\{x_t\}$ converges to
\begin{align}
q_\theta(x) = \frac{1}{Z(\theta)} \exp\{-E_\theta(x)\}
\end{align}
in distribution for an appropriate choice of $\{\eta_t\}$ \cite{welling2011,dalalyan2017}. We now consider the modified iteration
\begin{align}
x_{t + 1} = x_t - \frac{\alpha_t}{2} \nabla_x E_\theta(x) + \sqrt{\beta_t} \epsilon_t, \qquad t = 0, 1, 2, \ldots, T.
\end{align}
By the assumption $\alpha_t = \rho \beta_t$, the modified iteration is equivalent to
\begin{align}
x_{t + 1} &= x_t - \frac{\rho \beta_t}{2} \nabla_x E_\theta(x) + \sqrt{\beta_t} \epsilon_t, \qquad t = 0, 1, 2, \ldots, T \label{eq:prop2_2} \\
&= x_t - \frac{\beta_t}{2} \nabla_x (\rho E_\theta(x)) + \sqrt{\beta_t} \epsilon_t, \qquad t = 0, 1, 2, \ldots, T. \label{eq:prop2_3}
\end{align}
Since we have assumed the sequence generated by \eqref{eq:LMC_mod} converges, by comparing \eqref{eq:prop2_3} with \eqref{eq:prop2_1}, we conclude that the sequence $\{x_t\}$ generated by \eqref{eq:prop2_2} must converge to
\begin{align}
q_\theta^{\rho}(x) \coloneqq \frac{1}{Z(\theta, \rho)} \exp\left\{ - \rho E_\theta(x) \right\}
\end{align}
in distribution.
\end{proof}

\textbf{Proposition 3.} \textit{Assume the EBM $q_\theta$ is trained via MLE with convergent modified LMC \eqref{eq:LMC_mod} with $\alpha_t / \beta_t = \rho > 0$. Then, $\theta$ such that}
\begin{align*}
q_\theta(x) \propto p(x)^{1/\rho}.
\end{align*}
\textit{is a stationary point of MLE with gradient ascent.}

\begin{proof}
Let $\hat{q}_\theta$ be the distribution of $\hat{x}$ produced by running convergent modified LMC \eqref{eq:LMC_mod} with $\alpha_t / \beta_t = \rho > 0$ on some proposal sample $x \sim q_0$. Then, EBM gradient update with MLE with \eqref{eq:LMC_mod} becomes
\begin{align}
\theta \leftarrow \theta + \EE_{\hat{q}_\theta}[\nabla_\theta E_\theta(x)] - \EE_p[\nabla_\theta E_\theta(x)]
\end{align}
and thus stationarity is achieved when
\begin{align}
\hat{q}_\theta = p. \label{eq:prop3_1}
\end{align}
By Proposition \ref{prop:2}, we have
\begin{align*}
q_\theta^\rho = \hat{q}_\theta,
\end{align*}
so if $q_\theta(x) \propto p(x)^{1/\rho}$, \eqref{eq:prop3_1} is satisfied.
\end{proof}

\newpage

\section{Experiment Details} \label{append:exp}

\subsection{Experiments in Section \ref{sec:overestimate}}

To learn the one-dimensional mixture of Gaussians, we use EBMs whose energy function is the squared distance bewteen input and output of a multi-layer perceptron (MLP) with four layers, each with 512 hidden units and leaky-ReLU activations with negative slope $0.2$. For SRLMC, we set $T = 40$, $\alpha_t = 0.001$, and $\beta_t = 0.0001$. The replay buffer size is $50k$ and SRLMC chain reinitialization rate is $0.05$. For all methods, batch size is $1k$ and the optimizer is SGD with no momentum and learning rate $0.01$. Each EBM was trained for $5k$ iterations on a single GTX 1080 GPU.

\subsection{Experiments in Section \ref{sec:exp}}

\textbf{2D MoG.} MoG consists of six Gaussians, whose means are $(\cos \theta, \sin \theta)$ for $\theta \in \{n \pi / 3 : n = 0, 1, \ldots, 5\}$ and covariance matrices are $\sigma^2 \mathbf{I}$ for $\sigma = 0.1$. We use EBMs whose energy function is the output of a MLP with four layers, each with 512 hidden units and leaky-ReLU activations with negative slope $0.2$. For SRLMC, we set $T = 40$ and $\alpha_t = 0.001$, and $\beta_t = 0.0001$. The replay buffer size is $50k$, SRLMC chain reinitialization rate is $0.05$, and batch size is $1k$. For PS-USP, we set $n_m = 1$ and $n_r = 1$ so we can combine maximization and repulsion into a single iteration (if $u_i$ does not violate constraint, apply maximization, otherwise, apply repulsion), and $N = 50$. We set $\epsilon = 0.05$, $n = 5k$, $|\Lambda| = 1k$, and $n_s = 5k$. For all methods, the optimizer is SGD with no momentum and learning rate $0.001$. Each EBM was trained until convergence with a single GTX 1080 GPU.

\textbf{FMNIST.} The FMNIST dataset was scaled into the range $[-1,1]$. For SRLMC, we also added Gaussian noise of standard deviation $0.1$ following the recommendation of previous works \cite{nijkamp2019,grathwohl2020,yang2021}. We use EBMs whose energy function is the output of a CNN with three convolution layers followed by two fully-connected layers. We use the leaky-ReLU activation with negative slope $0.4$. Each convolution layer has number of filters $\in \{32, 64, 128\}$ with filter size $3$ and stride $1$. Each convolution layer activation is followed by an average pooling layer with kernel size $2$ and stride $2$. For SRLMC, we set $T = 20$ and $\alpha_t = 2.0$ and $\beta_t = 0.01$. The replay buffer size is $50k$, SRLMC chain reinitialization rate is $0.05$, and batch size is $125$. For PS-USP, we proceed similar to the case of 2D MoG. We set $n_m = 1$, $n_r = 1$, $N = 100$, $\epsilon = 10$, $n \in \{10k, 25k, 50k\}$, $|\Lambda| = 125$, and $n_s = 625$. For all methods, the optimizer is Adam with learning rate $0.001$. Each EBM was trained for at most $50$ epochs with a single GTX 1080 GPU. We choose models based on FPR95 on the OOD validation datasets of MNIST and KMNIST following Elflein et. al \cite{elflein2021}.

\end{document}